# Maintaining Journalistic Integrity in the Digital Age: A Comprehensive NLP Framework for Evaluating Online News Content

Ljubisa Bojic[1], Ph. D.
Senior Research Fellow
The Institute for Artificial Intelligence Research and Development of Serbia
University of Belgrade, Institute for Philosophy and Social Theory, Digital Society Lab

Nikola Prodanovic[2], Ph. D.
Research Fellow
Institute for Artificial Intelligence Research and Development of Serbia

Agariadne Dwinggo Samala[3], Ph. D.
Assistant Professor
Universitas Negeri Padang, Faculty of Engineering, West Sumatera, Indonesia

## Abstract

The rapid growth of online news platforms has led to an increased need for reliable methods to evaluate the quality and credibility of news articles. This paper proposes a comprehensive framework to analyze online news texts using natural language processing (NLP) techniques, particularly a language model specifically trained for this purpose, alongside other well-established NLP methods. The framework incorporates ten journalism standards—objectivity, balance and fairness, readability and clarity, sensationalism and clickbait, ethical considerations, public interest and value, source credibility, relevance and timeliness, factual accuracy, and attribution and transparency—to assess the quality of news articles. By establishing these standards, researchers, media organizations, and readers can better evaluate and understand the content they consume and produce. The proposed method has some limitations, such as potential difficulty in detecting subtle biases and the need for continuous updating of the language model to keep pace with evolving language patterns.

[1] Corresponding author;
Email address: ljubisa.bojic@ivi.ac.rs;
ORCID: 0000-0002-5371-7975
Address of correspondence: 1 Fruskogorska, Novi Sad, Serbia;
[2] Email address: nikola.prodanovic@ivi.ac.rs;
ORCID: 0000-0001-8698-9830
Address of correspondence: 1 Fruskogorska, Novi Sad, Serbia;
[3] Email address: agariadne@ft.unp.ac.id;
ORCID: 0000-0002-4425-0605
Address of correspondence: 482X+WQ West Air Tawar, Padang City, West Sumatra, Indonesia;

**Introduction**

In today's digital age, the consumption of news has increasingly shifted from traditional print and broadcast media to online platforms. While this shift has made news more accessible than ever before, it has also raised concerns about the quality, accuracy, and credibility of online news content (Tandoc et al., 2018). The digital news landscape has also given rise to several challenges and issues that threaten the integrity of journalism and the quality of information that people consume. These issues include the proliferation of fake news (Vosoughi et al., 2018), algorithmic manipulation, technology addiction, polarization (Bakshy et al., 2015; Pariser, 2011), and more.

Fake news has become a significant concern, particularly with the rise of social media platforms like Facebook and Twitter, which have become primary sources of news for many people (Allcott & Gentzkow, 2017). These platforms, with their algorithms optimized for engagement, have inadvertently created an environment where misinformation and sensationalized content can spread like wildfire (Vosoughi, Roy, & Aral, 2018). The prevalence of fake news not only undermines the credibility of journalism as a whole but also poses serious threats to democracy, as it can influence public opinion and affect political outcomes (Lazer et al., 2018).

Another issue is the increasing polarization of news consumption, where people tend to gravitate towards sources that confirm their pre-existing beliefs and biases (Bakshy, Messing, & Adamic, 2015). This echo chamber effect can result in a more polarized society and make it difficult for individuals to engage in meaningful discussions and find common ground on important issues (Sunstein, 2017). Furthermore, research has demonstrated that people's exposure to diverse perspectives can be limited by algorithms that personalize online content (Pariser, 2011). These algorithms, designed to maximize user engagement, can reinforce pre-existing beliefs and contribute to the echo chamber effect.

Technology addiction, particularly in the form of excessive smartphone and social media use, has been linked to negative outcomes such as sleep deprivation, stress, and reduced cognitive functioning (Wilcockson, Ellis, & Shaw, 2018). Additionally, the constant bombardment of information and the pressure to stay up-to-date with the latest news can lead to information overload, causing individuals to struggle with effectively processing and critically evaluating the content they consume (Eppler & Mengis, 2004).

Algorithmic manipulation has also become a concern, as sophisticated techniques are used to influence public opinion and behavior through the spread of targeted misinformation or the manipulation of search engine results (Howard & Kollanyi, 2016). This can result in biased information reaching the public, further compromising the credibility and integrity of journalism.

Given these challenges, developing tools and methodologies to evaluate and ensure the quality of online news is of utmost importance.

One such area of development is natural language processing (NLP), a subfield of artificial intelligence and linguistics that focuses on the interaction between computers and human language (Jurafsky & Martin, 2019). Advances in NLP have led to the creation of various tools and techniques that can be used to analyze, categorize, and evaluate the content of news articles (Graves, 2018; Lazer et al., 2018). However, while these tools have shown promise in addressing certain aspects of news quality, there is still a need for a comprehensive framework to evaluate online news content based on established journalistic standards (Lewandowsky et al., 2012).

Recent advances in the field of NLP sparked by idea of transformer neural networks (Vaswani et al., 2017) enabled revolutionary new level of natural language understanding by computers. The neural architectures based on transformer architecture are now able to very precisely construct mathematical graph-vector objects of human written text. Such mathematical objects are readily usable for precise numerical analysis (Devlin et al, 2019) such as classification or regression or further generation of text corresponding to input embeddings (Radford et al, 2018). These advances, opened ability to process text plausible fast and contextually accurate. Historically, such level of performance of NLP tools was not possible, due to the hardware limitations and due to the absence of proper machine learning architectures which rather processed text in more statistical manner by using static word embeddings that were not able to capture contextual meaning of words.

We argue that both presence and absence of extensive and balanced journalism standards, no matter how complex they are, are now possible to be recognized with transformer-based technologies with the strong supervision support by journal-ethical experts.

Journalism standards are essential for ensuring the quality and credibility of news content (Plaisance, 2013). Various studies have identified and analyzed different journalistic standards, such as objectivity (Schudson, 2001), balance and fairness (Entman, 1993), readability and clarity (Gunning, 1952), sensationalism and clickbait (Vettehen et al., 2008), ethical considerations (Ward, 2010), public interest and value (Curran, 2010), source credibility (Metzger et al., 2003), relevance and timeliness (Galtung & Ruge, 1965), and factual accuracy (Tandoc et al., 2018). These standards provide a foundation for evaluating the quality of news content across various platforms, including online news.

Historically, concerns about the quality and credibility of online news have led to increased interest in developing and applying NLP techniques for news analysis (Graves, 2018; Lazer et al., 2018). Recent studies have explored various NLP methods, such as sentiment analysis (Pang & Lee, 2008), topic modeling (Blei, 2012), and text classification (Sebastiani, 2002), to address specific aspects of news quality, such as objectivity, balance, and sensationalism (Conroy et al., 2015; Diakopoulos & Koliska, 2017). These approaches demonstrate the potential for NLP techniques to contribute to the evaluation of online news content.

However, while existing NLP techniques have shown promise in addressing certain aspects of news quality, there is still a need for a comprehensive framework that incorporates a broader range of journalism standards (Lewandowsky et al., 2012). Developing such a framework requires a synthesis of the existing literature on journalism standards, online news quality, and state of the art NLP techniques based on transformer neural networks, as well as an examination of potential gaps and opportunities for future research.

The noted literature review raises the research question: Can standards for NLP analysis be established based on existing literature to effectively evaluate the quality and credibility of online news texts? This paper seeks to answer this question by conducting a literature review on media and communication studies and examining the potential for developing a comprehensive framework for NLP analysis of online news content.

**Previous frameworks for evaluating news**

The growing presence of online news has sparked several attempts to develop frameworks for evaluating news quality and credibility. One approach involves using natural language processing (NLP) techniques to analyze textual features and detect biases in news content (Conroy, Rubin, & Chen, 2015). In a similar vein, Park, Khan, and Park (2018) implemented machine learning techniques to classify news articles based on their content and quality. These studies demonstrate the potential of NLP techniques in evaluating news content; however, they often focus on specific aspects or fall short of providing a comprehensive framework.

Several studies have attempted to develop criteria or guidelines for assessing journalistic quality. For instance, McQuail (2013) presented a set of media performance standards, such as objectivity, balance, and ethical considerations. While valuable, these criteria need to be operationalized in the context of digital news to be effectively applied. By incorporating these standards into a comprehensive NLP framework, as proposed in our paper, we aim to bridge this gap.

The use of transformer-based language models in text content analysis has also been explored in prior research. For example, Devlin, Chang, Lee, and Toutanova (2019) introduced the BERT (Encoder Transformer) model, which has been effectively used in various NLP tasks. However, the application of BERT and other language models to the analysis of news content remains limited partially due to text length limitations (Beltagy et al., 2020) and absence of labeled datasets by journalist-ethical experts. Our proposed framework addresses this limitation by incorporating a language model specifically trained for evaluating online news content by suitable expert supervision.

Previous research has identified specific issues related to online news, such as sensationalism and clickbait (Chakraborty et al., 2016). These studies underscore the need for a comprehensive approach that considers all aspects of news content, as our proposed framework does.

While existing literature has explored the use of NLP techniques in news analysis and established journalistic standards, there is a need for a comprehensive framework that integrates these aspects. Our study contributes to filling this gap by proposing a method that incorporates a language model and ten journalism standards to assess the quality of online news articles which will be further used for supervising language model.

**Framework of journalist standards**

*Objectivity*

Objectivity in the context of journalism and media refers to the fair, neutral, and unbiased presentation of news and information. It is a fundamental principle that guides journalists in their reporting and serves as a standard against which the quality and credibility of news are measured (Schudson, 2001).

Objectivity is closely related to the concept of empirical evidence and the scientific method. Empirical evidence is information gathered through observation, measurement, and experimentation, rather than subjective opinions or personal beliefs (Popper, 1959). The scientific method involves systematic observation, measurement, and experimentation, as well as the formulation, testing, and modification of hypotheses, to ensure the validity and reliability of research findings (Kuhn, 1962).

In journalism, objectivity can be characterized by several key components, such as neutrality, balance, impartiality, and accuracy (Tuchman, 1972). Neutrality refers to the absence of bias or personal opinions in news reporting, while balance involves the fair and equal representation of different perspectives and viewpoints. Impartiality requires journalists to remain detached from the subjects they cover, avoiding any real or perceived conflicts of interest that could compromise their reporting. Accuracy, on the other hand, involves the careful verification of facts and data to ensure the validity and trustworthiness of news content (Schudson, 2001).

The concept of objectivity has been widely debated and criticized in media and communication literature. Some scholars argue that true objectivity is unattainable, as journalists inevitably bring their own biases, values, and perspectives to their reporting (Gitlin, 1980). Others contend that objectivity can serve as a normative ideal or guiding principle, helping journalists to strive for fairness and impartiality in their work (Schudson, 2001).

Despite these ongoing debates, the pursuit of objectivity remains a crucial aspect of journalistic practice. In an era of information overload and widespread misinformation, the ability to distinguish between objective, fact-based reporting and subjective, opinion-driven content is more important than ever (Lewandowsky, Ecker, & Cook, 2017). By adhering to the principles of objectivity, journalists can help to promote informed public discourse and foster trust in the media.

*Balance and fairness*

Balance and fairness are essential components of journalistic objectivity and play a crucial role in ensuring the credibility and trustworthiness of news reporting (Entman, 1993). These concepts can be understood as the accurate and proportional representation of diverse perspectives, evidence, and stakeholders within a news story or media coverage (Donsbach & Klett, 1993).

Balance in journalism refers to the equitable presentation of different viewpoints, opinions, and arguments on a particular issue or topic. This may involve presenting opposing perspectives, ensuring that various stakeholders are represented, and providing a comprehensive overview of the available evidence (Strömbäck, 2008). Balance can be assessed by examining the range and diversity of sources, the proportionality of coverage, and the inclusion of different types of evidence and data within a news story (Donsbach & Klett, 1993).

Fairness, on the other hand, is closely related to the concept of impartiality and involves treating all parties and viewpoints with equal consideration and respect. This may include avoiding the use of biased language, refraining from favoring one side over another, and ensuring that all relevant information is presented in a clear and accurate manner (Entman, 1993). Fairness can be evaluated by examining the framing and tone of media coverage, the use of loaded language or emotive appeals, and the degree to which news reporting adheres to ethical guidelines and professional standards (Strömbäck, 2008).

Both balance and fairness have been the subject of extensive research and debate within media and communication studies. Some scholars argue that the pursuit of balance can lead to a phenomenon known as "false balance," wherein journalists give equal weight to opposing viewpoints, even when one side is supported by a preponderance of evidence (Boykoff & Boykoff, 2004). This can result in the distortion of scientific consensus and the misrepresentation of factual information (Koehler, 2016).

Despite these challenges, the principles of balance and fairness remain integral to the practice of journalism and the assessment of media quality. By striving to achieve balance and fairness in their reporting, journalists can help to foster informed public discourse, promote critical thinking, and enhance the credibility and trustworthiness of the news media.

*Readability and clarity*

Readability and clarity are essential aspects of effective communication in journalism. They refer to the ease with which a text can be read, understood, and processed by its intended audience (DuBay, 2004). Readability and clarity can be assessed using various metrics, linguistic analyses, and cognitive principles to ensure that information is presented in an accessible and comprehensible manner (Gunning, 1968).

Readability can be quantified using various formulas and indices, such as the Flesch-Kincaid readability index, Gunning Fog index, and the SMOG index, which consider factors such as sentence length, word complexity, and syllable count (Kincaid et al., 1975; Gunning, 1968; McLaughlin, 1969). These indices provide a numerical estimate of the reading difficulty of a text, allowing for comparisons across different texts and genres. However, it is important to note that readability metrics are only approximations and may not fully capture the nuances of language and comprehension (Benjamin, 2012).

Clarity, on the other hand, refers to the organization, structure, and presentation of information within a text. This may involve the use of clear and concise language, logical organization and coherence, and the avoidance of jargon and technical terms that may impede understanding (Alley, 2013). Clarity can be assessed by examining the flow and coherence of a text, the use of appropriate terminology and definitions, and the overall ease with which the content can be understood by a target audience (Plaxco, 2010).

Both readability and clarity are essential for ensuring effective communication, particularly in the context of journalism, where complex ideas and concepts must be conveyed to a diverse and often non-specialist audience (Fahnestock, 2011). By optimizing readability and clarity, journalists can enhance the accessibility of their content, promote public understanding of complex issues, and foster informed decision-making and public discourse (Bucchi & Trench, 2014).

*Sensationalism and clickbait*

Sensationalism and clickbait are journalistic practices that prioritize attention-grabbing headlines, provocative content, and emotional appeals over accurate, balanced, and objective reporting (Vettehen, Nuijten, & Beentjes, 2011). These practices can distort the presentation of news and information, contribute to misinformation, and undermine public trust in the media (Chen & Conroy, 2015).

Sensationalism refers to the use of exaggerated, dramatic, or emotional language and imagery to capture the audience's attention and elicit strong reactions (Grabe, Zhou, & Barnett, 2001). Sensationalism can be assessed by examining the degree to which news content relies on emotional appeals, hyperbole, and vivid descriptions, as well as the extent to which such content deviates from objective, fact-based reporting (Vettehen et al., 2011). Sensationalism can also manifest itself in the selection and framing of news stories, with media outlets prioritizing stories that are shocking, scandalous, or emotionally charged over more mundane or complex issues (Grabe et al., 2001).

Clickbait, on the other hand, is a more recent phenomenon that has emerged in the digital era of journalism. It refers to the practice of crafting headlines and article previews that are intentionally misleading, provocative, or sensational in order to entice readers to click on a link and drive web traffic (Blom & Hansen, 2015). Clickbait headlines often make use of curiosity gaps, cliffhangers, or exaggerated claims to pique the reader's interest and encourage them to engage with the content (Chen & Conroy, 2015).

Clickbait can be analyzed by examining the language and structure of headlines, as well as the relationship between the headline and the actual content of the article. This may involve assessing the use of provocative or misleading language, the presence of curiosity gaps, and the degree to which headlines accurately represent the information contained within the article (Blom & Hansen, 2015).

Both sensationalism and clickbait have been the subject of extensive research and criticism in media and communication studies. These practices have been linked to the decline of journalistic standards, the spread of misinformation and fake news, and the erosion of public trust in the media (Vettehen et al., 2011; Chen & Conroy, 2015). By understanding and addressing these issues, journalists and media organizations can work to promote more responsible, accurate, and objective reporting, ultimately contributing to a healthier media ecosystem and more informed public discourse.

*Ethical considerations*

Ethical considerations are fundamental to responsible journalism and media practices, as they guide journalists in making decisions that balance the public's right to information with the potential harm that may be caused by reporting (Ward, 2018). Ethical considerations can be understood as the principles, norms, and guidelines that inform the decision-making process and ensure that journalism adheres to professional standards and respects the rights and interests of all stakeholders involved (Plaisance, 2013).

Some of the key ethical considerations in journalism include truth-telling, accuracy, fairness, minimizing harm, respecting privacy, and avoiding conflicts of interest (Society of Professional Journalists, 2014). Truth-telling and accuracy require journalists to verify facts, correct errors, and ensure that their reporting is based on reliable and credible sources (Ward, 2018). Fairness involves presenting diverse perspectives, treating all parties with respect, and avoiding bias or discrimination in the reporting process (Plaisance, 2013).

Minimizing harm is a critical ethical principle that requires journalists to consider the potential consequences of their reporting on individuals, communities, and society at large. This may involve weighing the public interest against the potential harm caused by disclosing sensitive information, protecting vulnerable sources, or avoiding the dissemination of graphic or traumatic content (Ward, 2018).

Respecting privacy involves balancing the public's right to know with the individual's right to privacy, recognizing that some personal information may be off-limits or require special consideration before being made public (Society of Professional Journalists, 2014). Avoiding conflicts of interest requires journalists to maintain their professional independence, disclose any potential biases or financial interests, and refrain from accepting gifts or favors that may compromise their objectivity (Plaisance, 2013).

Ethical considerations can be assessed by examining the adherence of news reporting to established ethical guidelines, such as the Society of Professional Journalists' Code of Ethics or the International Federation of Journalists' Declaration of Principles. This may involve evaluating the accuracy and fairness of news coverage, the treatment of sources and subjects, the disclosure of conflicts of interest, and the overall quality and credibility of news content (Ward, 2018).

By incorporating ethical considerations into their reporting, journalists can contribute to the development of more responsible, trustworthy, and credible media, fostering public trust and promoting informed public discourse.

*Public interest and value*

Public interest and value are crucial considerations in journalism and media practices, as they guide the selection, presentation, and evaluation of news content based on its relevance, importance, and potential impact on society (Curran, 2010). Public interest and value can be understood as the degree to which news reporting contributes to public discourse, informs decision-making, and enhances the overall knowledge and understanding of important issues (Schudson, 2008).

Public interest refers to the topics, issues, and events that are deemed significant or relevant to a broad audience, often encompassing matters that affect the well-being, rights, or interests of society as a whole (Curran, 2010). This may include topics such as public health, safety, governance, social justice, and the environment, among others. Assessing public interest may involve evaluating the relevance of news content to its intended audience, the potential impact of the information on public opinion or policy, and the extent to which the content addresses pressing societal concerns (Schudson, 2008).

Value, on the other hand, refers to the overall quality, importance, and usefulness of news content in informing, engaging, and enlightening audiences (Carpini, 2000). This may encompass various dimensions of news reporting, such as accuracy, objectivity, depth, and timeliness, as well as aspects related to storytelling, narrative, and presentation. Assessing value may involve evaluating the contribution of news content to public knowledge, the effectiveness of the content in promoting understanding and engagement, and the potential for the content to inspire further inquiry or debate (Schudson, 2008).

Both public interest and value are essential for ensuring that journalism serves its core function as a provider of accurate, relevant, and meaningful information that fosters informed decision-making and public discourse (Carpini, 2000). By prioritizing public interest and value in their reporting, journalists can help to promote a more informed, engaged, and democratic society, ultimately contributing to the overall health and vitality of the media ecosystem.

*Source credibility*

Source credibility is a crucial factor in assessing the quality and trustworthiness of news content, as it refers to the perceived reliability, expertise, and trustworthiness of the sources cited in news articles (Metzger & Flanagin, 2015). Source credibility can be evaluated by examining the reputation, qualifications, and potential biases of the sources used in news reporting, as well as the degree to which these sources are accurately and transparently represented (Hovland & Weiss, 1951).

There are several dimensions of source credibility, including expertise, trustworthiness, and objectivity (Hovland, Janis, & Kelley, 1953). Expertise refers to the knowledge, skills, and competence of a source in a particular domain, which can be assessed based on their qualifications, experience, and track record (Fiske, Cuddy, Glick, & Xu, 2002). Trustworthiness involves the honesty, integrity, and reliability of a source, as well as their motivations and potential biases (Pornpitakpan, 2004). Objectivity requires that a source is perceived as impartial, unbiased, and free from external influences or conflicts of interest (Metzger & Flanagin, 2015).

In news reporting, the credibility of sources can be enhanced by providing clear and transparent attribution, verifying information through multiple sources, and ensuring that sources are appropriately qualified and relevant to the topic at hand (Tsfati & Ariely, 2014). Assessing source credibility may involve evaluating the reputation and expertise of the sources cited, the consistency and accuracy of their statements, and the potential for conflicts of interest or biases that may influence their opinions or findings (Hovland et al., 1953).

Source credibility is essential for ensuring the accuracy, reliability, and trustworthiness of news content, as well as for fostering public trust in the media (Tsfati & Ariely, 2014). By carefully selecting, verifying, and representing sources, journalists can contribute to the development of more responsible, credible, and informative news reporting, ultimately promoting informed public discourse and enhancing the overall quality of the media ecosystem.

*Relevance and timeliness*

Relevance and timeliness are important criteria for evaluating news content, as they ensure that the information provided is both pertinent to the target audience and current with respect to ongoing events and developments (Harcup & O'Neill, 2001). Relevance and timeliness can be assessed by examining the connection between news content and the interests, concerns, or needs of the audience, as well as the currency and immediacy of the information presented (Shoemaker & Vos, 2009).

Relevance refers to the degree to which news content is meaningful, significant, or applicable to a specific audience or context (Harcup & O'Neill, 2001). This may involve considering factors such as the geographic location, demographic characteristics, or cultural background of the audience, as well as the overall importance and impact of the information on their lives, values, or interests (Shoemaker & Vos, 2009). Assessing relevance may involve evaluating the connection between news content and the needs or concerns of its intended audience, the potential for the content to inform or engage readers, and the extent to which the content addresses pressing societal issues or challenges (Harcup & O'Neill, 2001).

Timeliness, on the other hand, refers to the currency and immediacy of news content in relation to ongoing events, developments, or trends (Shoemaker & Vos, 2009). This may involve considering factors such as the recency of the information, the speed with which it is disseminated, and the degree to which it is updated or revised in response to new developments (Harcup & O'Neill, 2001). Assessing timeliness may involve evaluating the currency of news content in relation to recent events or issues, the responsiveness of the content to emerging trends or developments, and the potential for the content to influence or inform public discourse and decision-making (Shoemaker & Vos, 2009).

Both relevance and timeliness are essential for ensuring that news content is informative, engaging, and responsive to the needs and interests of its audience (Harcup & O'Neill, 2001). By prioritizing relevance and timeliness in their reporting, journalists can help to foster informed public discourse, promote critical thinking, and enhance the overall quality and value of the news media.

*Factual accuracy*

Factual accuracy is a fundamental aspect of responsible journalism and media practices, as it ensures that the information presented to the public is reliable, trustworthy, and based on verifiable evidence (Graves, 2016). Factual accuracy can be assessed by examining the validity and reliability of the facts, data, and claims reported in news articles, as well as the degree to which these elements are supported by credible sources and evidence (Silverman, 2015).

Ensuring factual accuracy in news reporting involves several key practices, including verifying information through multiple sources, cross-referencing facts and data with authoritative references, and correcting errors and inaccuracies promptly and transparently (Kovach & Rosenstiel, 2014). Assessing factual accuracy may involve evaluating the consistency and coherence of the information presented, the credibility of the sources cited, and the potential for biases or distortions that may influence the interpretation or presentation of facts (Graves, 2016).

Factual accuracy is essential for maintaining the credibility and trustworthiness of the news media, as well as for fostering informed public discourse and decision-making (Lewandowsky, Ecker, & Cook, 2017). Inaccurate or misleading information can have serious consequences for public understanding and policy-making, particularly in areas such as public health, the environment, and social issues (Lewandowsky et al., 2017).

In recent years, the rise of misinformation, disinformation, and "fake news" has highlighted the importance of factual accuracy in journalism and media practices (Tandoc Jr, Lim, & Ling, 2018). By prioritizing factual accuracy in their reporting, journalists can contribute to the development of more responsible, credible, and informative news content, ultimately promoting public trust in the media and enhancing the overall quality of the media ecosystem.

*Attribution and transparency*

Attribution and transparency are essential components of responsible journalism and media practices, as they contribute to the credibility, trustworthiness, and accountability of news content (Karlsson, 2010). Attribution and transparency can be assessed by examining the clarity and accuracy with which sources, evidence, and information are presented, as well as the degree to which the journalistic process is open and accessible to scrutiny and evaluation (Bruns, 2008).

Attribution refers to the practice of crediting the sources of information, evidence, and claims used in news reporting, including the identification of individuals, organizations, or documents from which the information is derived (Karlsson, 2010). Proper attribution is crucial for ensuring the reliability and accuracy of news content, as it allows readers to evaluate the credibility and expertise of the sources cited and to verify the information independently if necessary (Tandoc, 2014). Assessing attribution may involve examining the clarity and accuracy with which sources are identified, the consistency of the information provided, and the degree to which the sources are appropriately credited and acknowledged (Karlsson, 2010).

Transparency, on the other hand, refers to the openness and accessibility of the journalistic process, including the methods, techniques, and decisions involved in gathering, selecting, and presenting news content (Bruns, 2008). This may involve providing clear explanations of the editorial process, disclosing potential conflicts of interest, or inviting input and feedback from readers and other stakeholders (Karlsson, 2010). Assessing transparency may involve examining the degree to which news reporting is open to scrutiny and evaluation, the extent to which editorial decisions and processes are explained and justified, and the potential for dialogue and interaction between journalists and their audience (Bruns, 2008).

By emphasizing attribution and transparency in their reporting, journalists can contribute to the development of more credible, trustworthy, and accountable news content, ultimately fostering public trust in the media and promoting informed public discourse (Karlsson, 2010).

**NLP analysis of online news content using journalistic standards**

*Applying Journalist standards*

Natural language processing (NLP) could become an essential tool in the age of digital journalism, where vast amounts of online news content are produced daily. By using NLP techniques, researchers and practitioners can automatically analyze and evaluate the quality of news articles, thus helping to maintain high journalistic standards and preserve the integrity of journalism. In this chapter, we will argue how the aforementioned journalistic standards can be used to develop a comprehensive framework for NLP analysis of online news content.

Objectivity is a crucial aspect of journalistic standards, as it ensures that news articles present information in a neutral and unbiased manner. NLP techniques such as sentiment analysis and opinion mining can be employed to detect subjective language, opinionated statements, or strong emotional expressions in news articles (Pang & Lee, 2008). By identifying these elements, an NLP framework can help assess the objectivity of a given article and flag articles that may be biased or slanted.

Ensuring balance and fairness in news articles involves presenting diverse perspectives and fairly representing different viewpoints. NLP techniques such as topic modeling and argument mining can be employed to analyze the representation of various stakeholders, the balance of quotes, and the overall framing of the issue (Greene & Cross, 2017). By assessing these elements, an NLP framework can help evaluate the extent to which an article adheres to the principles of balance and fairness.

An important aspect of journalistic standards is ensuring that news articles are easy to read and understand. NLP techniques such as text complexity analysis and lexical richness measures can be used to assess sentence complexity, jargon usage, and overall coherence in news articles (McNamara et al., 2014). By evaluating these elements, an NLP framework can help ensure that articles are accessible and comprehensible to a wide range of readers.

In the digital age, sensationalism and clickbait have become pervasive in online news content. NLP techniques such as headline analysis and language modeling can be used to detect exaggerated claims, provocative language, or attention-grabbing headlines in news articles (Chakraborty et al., 2016). By identifying these elements, an NLP framework can help combat the spread of sensationalist and misleading content in online journalism.

Ethical considerations are fundamental to maintaining high journalistic standards. NLP techniques such as named entity recognition and relationship extraction can be employed to examine articles for potential ethical concerns, such as invasion of privacy, harm to individuals or groups, or conflicts of interest (Manning & Schütze, 1999). By evaluating these elements, an NLP framework can help ensure that articles adhere to journalistic standards and ethical guidelines.

Assessing the public interest and value of news articles involves evaluating their significance, impact, and potential for informing or engaging readers. NLP techniques such as text summarization and topic modeling can be employed to analyze the content and focus of news articles, while social media analytics can be used to gauge reader engagement and dissemination (Agarwal et al., 2018). By assessing these elements, an NLP framework can help determine the overall value and contribution of news articles to public discourse and knowledge.

Source credibility is a vital aspect of journalistic standards, as it ensures that the information presented in news articles is reliable and trustworthy. NLP techniques such as citation analysis and stance detection can be employed to assess the credibility of the sources cited in news articles, such as experts, eyewitnesses, or documents (Dori-Hacohen & Allan, 2015). By evaluating these elements, an NLP framework can help maintain the integrity of journalism by ensuring that articles rely on credible and unbiased sources.

An important criterion for news articles is their relevance to the intended audience and their timeliness in relation to current events or issues. NLP techniques such as event extraction and temporal analysis can be employed to analyze the focus, context, and connection of news articles to recent developments (Mazur & Dale, 2017). By assessing these elements, an NLP framework can help ensure that articles are relevant, timely, and engaging for readers.

Factual accuracy is a cornerstone of journalistic standards, as it ensures that news articles present reliable and accurate information. NLP techniques such as fact-checking, data validation, and misinformation detection can be employed to cross-reference facts or data with credible sources and identify false or misleading information in news articles (Shu et al., 2017). By evaluating these elements, an NLP framework can help maintain the credibility and trustworthiness of journalism in the digital age.

Attribution and transparency play a crucial role in journalistic standards, as they ensure that news articles properly acknowledge the sources of information, quotes, and data. NLP techniques such as text similarity measures, citation analysis, and named entity recognition can be employed to identify instances of plagiarism, misattribution, or lack of proper citations in news articles (Osman et al., 2018). By evaluating these elements, an NLP framework can help maintain the credibility and integrity of journalism by ensuring that articles are transparent and accountable in their use of sources and information.

By incorporating these journalistic standards into a comprehensive framework for NLP analysis of online news content, researchers and practitioners can effectively assess the quality of news articles and maintain high journalistic standards in the digital age. This framework can serve as a valuable tool for news organizations, regulators, and readers alike, ensuring that journalism remains a vital and trustworthy source of information in the digital era.

*Enhancing the NLP Framework for Evaluating Online News Content*

In order to address the limitations and enhance the effectiveness of the proposed comprehensive framework for evaluating online news content using NLP techniques, several methods can be considered.

One method involves employing advanced sentiment analysis techniques, such as aspect-based sentiment analysis (ABSA) or deep learning models, to better detect subtle biases in the language used within news articles (Pontiki et al., 2016). These techniques can help identify opinionated expressions related to specific aspects or entities, leading to a more fine-grained assessment of objectivity and balance in news content.

Another approach is to utilize state-of-the-art contextual language models based on transformer architecture, such as BERT, GPT-3, or RoBERTa, to improve the detection of subtle linguistic patterns and biases (Devlin et al., 2019; Radford et al., 2019; Liu et al., 2019). These pre-trained models have demonstrated strong performance in a variety of NLP tasks and can be fine-tuned for specific tasks related to journalistic standards evaluation, such as sentiment analysis or text classification.

Domain adaptation techniques can also be applied to transfer the knowledge gained from analyzing one type of news content to another (Pan & Yang, 2010). By leveraging transfer learning and domain adaptation, the NLP framework can be adapted to different news domains, increasing its generalizability and applicability across various news platforms and genres.

The framework can be extended to support multilingual and cross-cultural analysis of news content (Ruder et al., 2019). By incorporating NLP techniques that can handle multiple languages and cultural contexts, the framework can be applicable to a broader range of news sources, fostering a more inclusive and global evaluation of journalistic standards.

In addition, the analysis of visual and multimedia content, such as images, videos, and infographics, can be incorporated to evaluate the overall quality and credibility of news articles (Wang et al., 2019). Multimedia content can also convey biases, exaggerations, or misinformation. Combining NLP techniques with computer vision and multimedia analysis can provide a more comprehensive assessment of online news content. This is also the point where transformer architecture may contribute substantially as the application of the architecture has shown great promise in text-image multi-modality processing (Dosovitskiy et al., 2021).

It is essential to regularly update the language models and NLP techniques used in the framework to keep pace with evolving language patterns and emerging news content (Raffel et al., 2020). This can be achieved through periodic retraining of the models on new data and incorporating the latest advances in NLP research.

Finally, integrating a human-in-the-loop approach, where subject matter experts or journalists provide feedback and annotations to the NLP framework, can help improve the model's performance, address potential blind spots, and ensure that the framework remains aligned with the nuances and complexities of journalistic standards (Holgate et al., 2018).

**Conclusion**

This paper aimed to address the research question: Can standards for NLP analysis be established based on existing literature to effectively evaluate the quality and credibility of online news texts? By conducting a literature review on media and communication studies, journalism, and NLP, the study has identified ten journalism standards that can serve as the foundation for a comprehensive framework for NLP analysis of online news texts. These standards include objectivity, balance and fairness, readability and clarity, sensationalism and clickbait, ethical considerations, public interest and value, source credibility, relevance and timeliness, factual accuracy, and attribution and transparency.

The incorporation of these journalism standards into NLP tools and techniques offers a promising approach to evaluating online news content effectively. By establishing standards for NLP analysis, researchers, media organizations, and readers can better assess and understand the quality of news articles in the digital age. However, developing a comprehensive framework for NLP analysis of online news requires a synthesis of the existing literature on journalism standards, online news quality, and NLP techniques, as well as an examination of potential gaps and opportunities for future research.

The findings of this study provide a partial answer to the research question, demonstrating that standards for NLP analysis can indeed be established based on existing literature. However, there are limitations to this study, as the literature review may not have captured all relevant sources or perspectives, and the proposed framework may require refinement and validation through empirical testing. Additionally, the development of NLP tools and techniques that incorporate these journalism standards is still an ongoing process, and further research is needed to fully realize the potential of this approach. Developing and implementing such a framework can empower individuals to become more discerning consumers of news, promote critical thinking, and encourage informed decision-making. Furthermore, by holding news sources accountable to these journalism standards, the proposed framework can also contribute to the restoration of trust in journalism as a whole.

Future research should focus on expanding and refining the proposed framework, as well as developing and testing specific NLP tools and techniques that incorporate these journalism standards. For instance, researchers may consider exploring innovative techniques for detecting and measuring sensationalism and clickbait in news headlines, or for assessing the balance and fairness of news articles by analyzing the representation of various stakeholders and viewpoints. Moreover, interdisciplinary collaboration between media and communication studies, journalism, and NLP can provide valuable insights and contribute to the further enhancement of the evaluation of online news content.

Another area for future research is the examination of the ethical implications of using NLP techniques to evaluate online news content. As NLP tools become more sophisticated and ubiquitous, questions regarding privacy, surveillance, and the potential for misuse of these technologies will inevitably arise. Researchers should engage in critical discussions and consider the development of ethical guidelines for the use of NLP technologies in the context of news analysis.

Finally, given the global nature of online news consumption and production, future research should also explore the applicability of the proposed framework and NLP techniques across different cultural and linguistic contexts. By identifying and synthesizing key journalism standards and examining their potential for incorporation into NLP tools and techniques, this study contributes to the ongoing efforts to ensure the quality and credibility of news content in the digital age.